\documentclass[10pt,journal]{IEEEtran}
\usepackage{textcomp}
\usepackage[pdftex]{graphicx}

\graphicspath{{diagrams/}}
\DeclareGraphicsExtensions{.pdf,.jpeg,.png,.eps}
\usepackage{epstopdf}
\usepackage{algorithm2e} 
\usepackage{algpseudocode} 
\usepackage[frenchb, english]{babel}
\usepackage{csquotes}
\usepackage{cite}
\usepackage{caption}
\usepackage{subcaption}
\usepackage[cmex10]{amsmath}
\usepackage{float,multirow}
\usepackage{amsfonts}
\usepackage[none]{hyphenat}
\usepackage{url}

\usepackage{breakurl}
\usepackage[breaklinks]{hyperref}
\usepackage{soul}
\usepackage{dirtytalk}
\usepackage{hyperref}
\usepackage{comment}
\usepackage[euler]{textgreek}
\usepackage{epsfig,bm,amsmath,amssymb,graphicx,setspace}
\usepackage{color}
\usepackage[T1]{fontenc}
\usepackage{algpseudocode}
\usepackage{textgreek}
\usepackage{amsthm}
\usepackage{array}
\usepackage{balance}
\usepackage{enumerate}
\usepackage[utf8]{inputenc}
\usepackage{array}
\usepackage{wrapfig}
\usepackage{multirow}
\usepackage{tabularx}
\usepackage[section]{placeins}
\usepackage{graphicx}
\usepackage{subcaption}
\usepackage[utf8]{inputenc}
\usepackage[export]{adjustbox}
\usepackage{wrapfig}
\usepackage{array}
\usepackage{multirow}
\usepackage{graphicx}
\usepackage[table,xcdraw]{xcolor}

\def\BibTeX{{\rm B\kern-.05em{\sc i\kern-.025em b}\kern-.08em
        T\kern-.1667em\lower.7ex\hbox{E}\kern-.125emX}}
\usepackage[T1]{fontenc}
\usepackage{multirow}
\usepackage[frenchb]{babel}
\newcolumntype{C}[1]{>{\centering\arraybackslash}m{#1}}
\newcolumntype{P}[1]{>{\centering\arraybackslash}p{#1}}

\begin{document}
  \title{\textcolor{black}{Quantum Artificial Intelligence for Mission-Critical Systems: Foundations, Architectural Elements, and Future Directions}}

\author{Siva Sai, Rajkumar Buyya~\IEEEmembership{Fellow,~ACM; Fellow,~IEEE }

\thanks{Siva Sai and Rajkumar Buyya are with the Quantum Cloud Computing and Distributed Systems (qCLOUDS) Laboratory, School of Computing and Information Systems, The University of Melbourne, Australia (e-mail: \{sivasaireddy.naga, rbuyya\}@unimelb.edu.au).}
}

\maketitle

\begin{abstract}
Mission critical (MC) applications such as defense operations, energy management, cybersecurity, and aerospace control require reliable, deterministic, and low-latency decision making under uncertainty. Although the classical Artificial Intelligence (AI) approaches are effective, they often struggle to meet the stringent constraints of robustness, timing, explainability, and safety in the MC domains. Quantum Artificial Intelligence (QAI), the fusion of artificial intelligence and quantum computing (QC), can potentially provide transformative solutions to the challenges faced by classical ML models. \textcolor{black}{QAI is a broader umbrella than Quantum Machine Learning (QML) and additionally includes quantum optimization, search, and reasoning; we use QAI throughout the paper for the field at large, and QML only for learning-specific subroutines.}
\textcolor{black}{The principal contributions of this work are: (i) \textcolor{black}{a} systematic survey of QAI methods analyzed through the lens of MC requirements like certification, robustness, and timing; (ii) a conceptual quantum cloud resource management and scheduling framework with deployment assumptions, complexity analysis, and failure-mode discussion; and (iii) an identification of the gaps between current QAI capabilities and MC systems requirements.}
We also propose a \textcolor{black}{conceptual} model for management of quantum resources and scheduling of applications driven by timeliness constraints. We discuss multiple challenges, including trainability limits, data access, and loading bottlenecks, verification of quantum components, and adversarial QAI. Finally, we outline future research directions toward achieving interpretable, scalable, and hardware-feasible QAI models for MC application deployment.
\end{abstract}

\begin{IEEEkeywords}
  Quantum computing, Quantum artificial intelligence, Mission critical systems, Aerospace, Defense, Disaster management, Cybersecurity, Energy and grid management, Technical foundations
\end{IEEEkeywords}

\section{Introduction}
Mission critical (MC) applications are those that must not fail because failure can lead to severe harm, injury, loss of life, large economic damage, or major outages \cite{avizienis2004basic}. An MC system aim to build a structured safety case that traces hazards, requirements, designs, and tests, so that acceptable risk is argued with evidence rather than based on assumption. An air traffic control system, where even a slight malfunction can endanger hundreds of lives is an example of an MC application. \textcolor{black}{The industrial control systems in nuclear plants and power grids are also mission critical applications, since their failure could cause catastrophic accidents or large-scale blackouts. Machine Learning (ML) has become a default approach for perception, decision support, and prediction, with its applications ranging from sensor monitoring to triaging incidents.} Classical ML uses hand-crafted features with models like Support Vector Machines (SVMs) and gradient boosting, while Deep Learning (DL) learns features directly from the data. This capability of DL enabled breakthroughs in vision, speech, and control. 
Both ML and DL-based systems have a significant number of applications in MC systems, but these gains are not without the weaknesses that are critical in MC settings. 
The models might be overconfident in out-of-distribution (OOD) inputs, which is risky in early-warning systems and anomaly screens. Research \cite{chakraborty2021survey} also showed that carefully altered adversarial data can fool the deep learning models, which is an integrity concern for autonomy stacks and safety monitors. The recent DL models are data-hungry and computationally heavy, and hence, acquiring labels for rare failures can be very expensive. \textcolor{black}{Quantum Machine Learning (QML) models have the potential to address some of these limitations in MC applications, particularly for specific tasks where quantum feature spaces or optimization offer advantages over classical approaches.}

\textcolor{black}{Quantum Artificial Intelligence (QAI) is an umbrella term for the intersection of quantum computing and artificial intelligence. It is important to note that QAI is a much broader field than QML. QAI encompasses quantum optimization, quantum search, quantum reasoning, fuzzy quantum logic, among others \cite{acampora2026quantum, baioletti2025quantum}. The scope of this paper is particularly limited to QAI methods applied to mission-critical systems, and we use QAI when referring to the field at large.} 
QAI uses a quantum processor for particular steps in the learning process, like data preparation, measurement, and expectation estimation. The remaining parts of the hybrid pipelines, like data handling, logging, optimization, and control, remain classical. The key idea behind QAI is that by representing a classical input in quantum states, the device would be able to estimate overlaps that can act as useful kernels for standard ML classifiers. QAI leverages quantum phenomena like superposition and entanglement for processing vast amounts of information simultaneously, allowing better exploration of complex solution spaces and faster computation. MC tasks often face imbalanced, scarce labels (near-misses, rare faults) and are supposed to operate with tight assurance (bounded behavior, traceability). QAI can potentially help with these cases in concrete ways. QAI models can potentially act as powerful feature-map engines that enhance the separability of features even with limited examples. With Quantum AI models, certain learning tasks require fewer experiments, particularly when the data comes from instruments (calibration, sensing, and characterization). Many mission critical tasks reduce to reconfiguration or hard scheduling with strict latency budgets. Variational quantum optimizers like Quantum Approximate Optimization Algorithm (QAOA) provide fixed-depth circuits that balance runtime and solution quality. 
When training in MC systems that must cross organizational boundaries, QAI can make use of QC techniques like quantum cryptography and blind quantum computing to protect queries, data, and model updates.

\textcolor{black}{To the best of our knowledge, this \textcolor{black}{work offers one of the early systematic syntheses of} QAI methods for mission-critical systems. We also propose \textcolor{black}{a} resource management framework specifically in the context of mission-critical systems.}
\textcolor{black}{The primary purpose of this work is to synthesize existing QAI for mission-critical systems research. In addition, we also offer a conceptual architectural contribution through the quantum resource management framework. The paper is conceptual and architectural in scope and we do not include original experiments or benchmarks; we have identified empirical validation on emerging quantum cloud platforms as a future direction.}
The rest of the paper is organized as follows: In Section \ref{bkgd}, we provide a brief background of quantum computing and mission critical systems. Section \ref{qml} elucidates quantum machine learning briefly from multiple perspectives. In Section \ref{techf}, we provide core mechanisms and algorithmic principles of QAI in MC systems, followed by applications in Section \ref{appns}. 
Section \ref{industry} provides an exploration of the positioning of QAI for MC systems in the industry in terms of deployment. 
In Section \ref{framework}, we propose a framework to handle the unique properties and constraints of quantum computing in resource management and scheduling systems. 
We outline several challenges and future directions in Section \ref{chall}, and finally conclude the paper in Section \ref{concl}.

\section{Background}\label{bkgd}

\subsection{Quantum Computing}
Quantum computing deals with algorithms and systems that manipulate information in quantum states, which exploit two underlying principles of quantum physics - superposition and entanglement in order to transform the probability amplitudes in ways that differ from classical bit operations. Noisy Intermediate Scale Quantum (NISQ) view is a widely used perspective in today's quantum hardware, where devices with tens-hundreds of qubits run non-trivial circuits, but limited coherence and gate noise limit the reliability and circuit depth. This framing stresses the role of hybrid quantum-classical workflows and sets realistic expectations for near-term applications. 

\textbf{Quantum Models}: There are two models that are widely used to think about quantum computers from application composition perfective: gate model and adiabatic/quantum annealing. The gate model builds a program from a small set of gates (primitive operations), analogous to logic gates in classical chips. DiVincenzo et al. \cite{divincenzo2000physical} set out a classical checklist of what hardware quantum hardware must provide - universal gates, scalable qubits, long stability, the ability to measure, and reliable ways to move information. In adiabatic or quantum annealing model, the problem is encoded in the energy landscape of a quantum system and guided towards a low-energy (good) solution. Research shows that this formulation is computationally equivalent to the gate model under ideal conditions \cite{aharonov2017interactive}.

Since the quantum states are fragile, large programs demand fault tolerance. In this connection, the threshold theorem states that if the individual operations are accurate enough, the error-correcting codes can suppress errors, thus allowing the running of very long computations also. Among many codes, the surface codes \cite{fowler2012surface} are prominent as they work on 2-D grids with only local connections and have relatively high error thresholds. This makes them appealing for future scale systems.

\begin{figure*}[]
    \centering
    \includegraphics[width=0.8\textwidth]{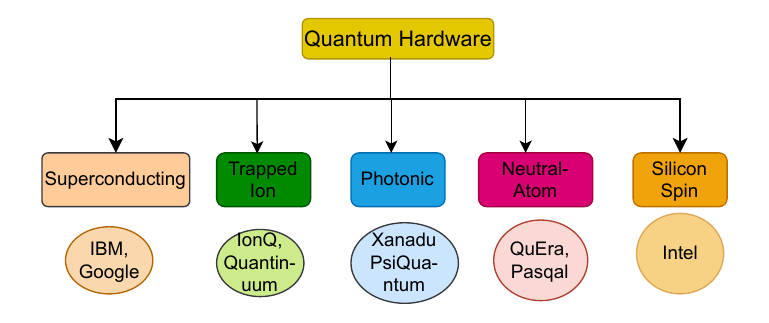}
    \caption{Leading approaches for quantum hardware realization and key industry players}
    \label{fig:qhardware}
\end{figure*}

\textbf{Quantum Architectures}: There are five leading approaches to building today's quantum hardware (refer Figure \ref{fig:qhardware}), and all of them have tradeoffs in terms of scale, control, and connectivity. Superconducting circuits \cite{kjaergaard2020superconducting} are based on microwave chips and enable fast gates and easy on-chip working, but have the issues of scaling and noise. Trapped ions \cite{bruzewicz2019trapped} achieve high fidelity by controlling individual atoms with lasers. A drawback of this approach is that the larger systems can be complex and slow to route. Natural atoms/Rydberg arrays \cite{saffman2016quantum} achieve scalable layouts with strong and switchable interactions by arranging many atoms in optimized tweezers. Single photons and integrated optics are used in Photonic platforms \cite{flamini2018photonic}, and they promise measurement-based logic and room-temperature operation. Semiconductor spin qubits \cite{zwanenburg2013silicon} leverage chip fabrication and store information in electron/hole spins in silicon quantum dots, but they are still maturing in control at scale. \textcolor{black}{ Apart from the above solid-state and photonic platforms, chemical pathways in the liquid phase have been explored where the fuzzy logic is processed using thermalized quantum mixed states \cite{gentili2025exploring}. Although this unconventional architecture remains at an early stage, it broaden the landscape of physical substrates for quantum information processing.}

Given the current limitations, the most practical quantum workflows are hybrid wherein the quantum processor executes a bounded task it is good at, like estimation overlaps or expectations, or preparing special states, while a classical processor handles the rest of the processes - data ingestion, optimization, and decision logic. Recent research reveals emphasis on the split in the tasks, the need to report noise levels, the number of measurements, and resource documents so that the results are reproducible and comparable across devices. 

\subsection{Mission Critical Systems}
The mission critical systems are the ones that must not fail because failure can cause large economic losses, major service outages, and injury or death. A few clear examples include flight control, power-grid protection, and railway signaling. The required aspects of an MC system include reliability (keeps working), safety (no unacceptable harm), integrity (no improper changes), availability (ready when needed), and maintainability (can be fixed quickly) \cite{avizienis2004basic}. In the MC systems, faults (root causes) are differentiated from errors (incorrect internal state) and failures (outwardly visible problems).

Timing is a defining constraint in MC systems - the systems have to respond within guaranteed deadlines. Classical real-time scheduling theory allows assigning priorities so that the periodic tasks meet their deadlines on a processor, the theory of which is still used in automotive controllers, industrial automation, and avionics. In MC systems, the timing analyses are paired with monitoring and redundancy so that the system continues operating safely even if the components of the system misbehave. Various standards are set forth in terms of the mission critical systems to ensure the required aspects in the systems. IEC 61508 defines safety integration levels and a safety lifecycle that scale the development rigor and risk testing. ISO 26262 is adopted in automotive systems and defines automotive safety integrity levels with requirements on software, hardware, and validation. SO-178C standard specifies design assurance levels in aviation, which are linked to failure severity, and also mandates strong testability based on requirements through tests.  

Across all MC domains, a common thread is that the hazard analysis must be done earlier, traceability from hazards must be maintained, an assurance level matching risk must be chosen, and investment in verification and validation must be made. So, for any new technology like quantum AI, acceptance in MC systems requires fitting into this framework.
\textcolor{black}{
To integrate the QAI components into these certification workflows, they should be treated as software items subject to the same assurance level the system supports. Under IEC 61508, this will imply assigning a Safety Integrity Level (SIL) to the QAI component based on the hazard it mitigates and providing evidence of system capability (testing coverage, design reviews) at that level. The QAI component would require a Design Assurance Level (DAL) with traceability from requirements to test cases under DO-178C. It implies, in practice, that the quantum circuits be version controlled, i.e, their transpiled outputs be verified against reference implementations with equivalence checking tools such as QCEC (Quantum Circuit Equivalence Checking) \cite{burgholzer2021qcec}. The stochastic nature of quantum measurement should also be bounded through repeated-shot statistical guarantees documented in the safety case.
}

\section{Quantum Machine Learning at a Glance} \label{qml}
\begin{figure*}
    \centering
    \includegraphics[width=0.8\textwidth]{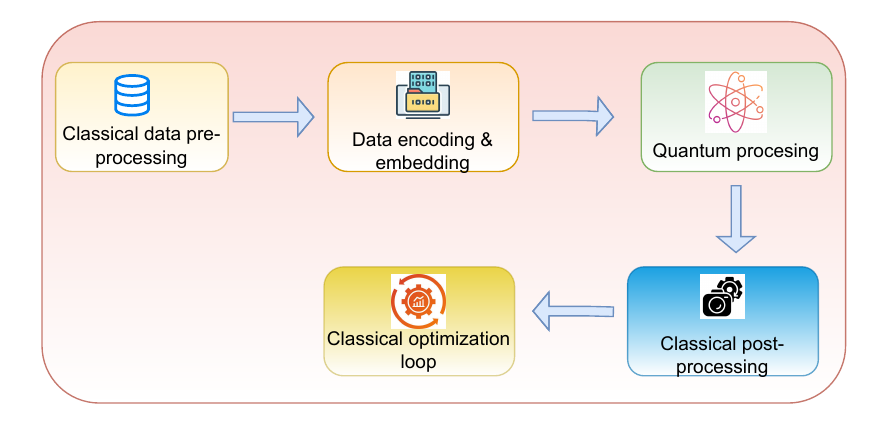}
    \caption{Workflow of a hybrid quantum-classical machine learning framework}
    \label{fig:qmlworkflow}
\end{figure*}

Quantum machine learning focuses on learning methodologies that use quantum information techniques at important steps like data representation, similarity measurement, or model optimization. Early reviews on QML helped in shaping the field, while cautioning against over-claiming. A well-cited early review on QML by Biamonte et al. \cite{biamonte2017quantum} placed QML as its own field (in the year 2017) and noted that any practical gains would depend on proper assumptions on hardware and data access. In 2019, the work by Schuld et al. \cite{schuld2019quantum} revealed a central idea in the field - by preparing the classical data as quantum states, the quantum devices can realize kernels that are associated with high-dimensional feature spaces (which are very costly to evaluate classically). These kernels could act as powerful inputs to the standard classifiers. Later on, researchers tested and verified this idea on actual hardware by developing an end-to-end workflow. Havlivcek et al. \cite{havlivcek2019supervised} built a quantum kernel estimator that feeds a variational quantum classifier and a classical SVM on a superconducting processor. A recent research by Huang et al. \cite{huang2022quantum} showed through demonstration on today's quantum hardware that a quantum approach can significantly reduce the number of physical experiments needed for particular \enquote{learning from experiments} tasks. Figure \ref{fig:qmlworkflow} shows the workflow of a hybrid QML framework. The workflow begins with preprocessing and encoding of the classical data for execution on a quantum circuit. The output of the quantum circuit is measured and post-processed classically and the output is fed into a classical optimization loop that updates the quantum circuit's parameters iteratively to train the model. 

Broadly, the QML models can be divided into four categories - supervised quantum kernel methods, variational or discriminative models, quantum generative models, and quantum reinforcement learning models. Quantum-kernel methods embed classical data \(x\) into quantum states \(|\phi(x)\textrangle\) and rely on the similarities that are estimated as inner products (kernels) by the model. The quantum kernel methods make use of a quantum processor only to compute the kernel matrix \(K_{ij} = |\textlangle\phi(x_i)|\phi(x_j)
\textrangle|^2\) and a classical model is used to handle the training. Variants of quantum-kernel methods differ in the feature map (amplitude, angle, or problem-structured encodings) and in terms of how the kernels are estimated. With these methods, the training remains a convex classical optimization, and hence they are attractive when the data are scarce and the decision boundaries are subtle. They also provide better audit trails (margins and support vectors). 
Variational or discriminative models (parametrized quantum circuits, PQCs) have their quantum part as a trainable circuit \(U(\theta)\) whose outputs (bitstrings, expectation values) are fed to a loss that is minimized by a classical optimizer. Cerezo et al. \cite{cerezo2021variational} provided a comprehensive review on variational quantum circuits, explaining the standard loop of prepare-measure-update, known training pathologies, design choices for ansätze, and mitigation strategies for near-term devices. Geometry-aware optimization and data re-uploading mechanisms increase the practical expressivity of PQCs without the need for deep circuits. PQCs are preferred in practicality when full end-to-end quantum training of the model is desired (instead of a \enquote{kernel-only} role). 
Quantum generative models help to learn data distributions using quantum processes \cite{saigenerative}. Quantum Generative Adversarial Networks (QGANs) train a quantum generator against a quantum or classical discriminator. Quantum Boltzmann machines (QBMs) model thermal-like distributions. Lloyd et al. \cite{lloyd2018quantum} present an information-theoretic perspective on quantum generative modeling. Generally, quantum generative models are utilized for tasks where sampling structure (interference effects) is central. They are also used in places where classical scoring can be integrated with the evaluations of sample quality or model likelihoods. 

Quantum reinforcement learning (QRL) uses quantum components inside policy/value approximators or uses quantum subroutines within the normal loop of reinforcement learning. Jerbi et al. \cite{jerbi2021quantum} present a survey covering algorithm templates (PQC  policies, exploration primitives, value estimation), small-scale demonstrations, and open issues. The survey focuses on where quantum elements plausibly fit and where the QRL models might be better than the variational classifiers or kernel methods. Across all the above categories, the practical QML models remain hybrid - quantum hardware is used for well-defined subroutines like kernel estimation and circuit evaluations, while data handling and optimization remain classical. 

QML models have multi-fold advantages compared to their classical counterparts. As discussed before, quantum kernel mapping makes the feature space more expressive, which is hard to achieve with classical methods. PQCs follow analytic gradient rules, which enable the estimation of exact derivatives on real devices without the need to resort to fragile finite differences. Schuld et al. \cite{schuld2019evaluating} showed that this makes gradient-based training practical on hardware and helps stabilize optimization pipelines. Some quantum architectures encode symmetry and locality directly into the model, benefiting from built-in inductive bias from symmetry and physics. For example, Quantum Convolutional Neural Networks (QCNNs) use shallow and hierarchical circuits that reflect renormalization ideas \cite{cong2019quantum}. QCNNs are proven to classify cases of matter with logarithmic complexity and good noise tolerance. Apart from the raw accuracy, the theory for generalization that represents how well a trained model will perform on new data is an emerging research direction in QML. They utilize quantification techniques like Fisher information/effective dimension, giving insights into when a quantum model is likely to generalize well or overfit \cite{banchi2021generalization}.  

Being a relatively newer field, QML is not without significant challenges that need to be overcome to realize its full range of benefits. The practical guidance in the current variational circuits literature is to prefer problem-structured and shallow circuits with careful initialization and optimization. This is due to the barren plateau problem, where large and unstructured PQCs have gradients that vanish as systems get larger. This makes training ineffective \cite{cerezo2021variational}.
Noise and resource overheads are another notable challenge in current-day QML models. Error mitigation can reduce bias for short circuits, but at the cost of extra sampling \cite{temme2017error}. Recent research showed that for a particular set of quantum linear-algebra-based learners,  quantum-inspired classical models achieve similar performance to that of quantum models under realistic input models. This denotes the experimentation in the lines of dequantization. Further, the comparisons with classical baselines can be skewed due to weak classical baselines and inconsistent processing. Standard datasets/splits, full reporting of noise and hardware, and matched feature budgets could be some of the solutions for the above-mentioned issue.

\textcolor{black}{It is also important to acknowledge that QML has not consistently outperformed classical methods, as of the current state of QML research. The dequantization results by Tang et al. \cite{tang2019quantum} showed that quantum-inspired classical algorithms can match the runtime advantages that were previously attributed to linear-algebra and quantum recommendation-based learners. Huang et al. \cite{huang2022quantum} demonstrated that the classical ML models on the same data achieve comparable accuracy to QML models, with the quantum advantages being limited to specific settings where data comes from quantum processes. It has also been noted that the noise-induced performance degradation has resulted in quantum classifiers underperforming optimized classical baselines on several benchmarks on current NISQ hardware \cite{khanal2023evaluating}. These findings indicate that the claimed QML advantages are task-dependent and conditional on hardware maturity.}

\section{Core Mechanisms and Algorithmic Principles}\label{techf}

\subsection{Quantum-enhanced Learning Pipelines}
Quantum-enhanced learning pipelines fit into today's control decision systems, where they can be used only when they are required for certain subtasks. They can integrate seamlessly into the current MC control and decision systems, while leveraging quantum processors only where they provide significant representational or computational advantages. The mission critical domains share common requirements - extracting meaningful insights and patterns from small \& high-dimensional datasets, maintaining predictable and stable model training under hardware constraints, and achieving required expressivity with shallow circuits. In this section, we try to map these requirements to quantum machine learning methods. 

In applications such as satellite telemetry anomaly detection and propulsion systems fault diagnosis, the classical ML models face limitations in handling limited samples of high-dimensional data. Quantum feature mapping projects the classical data into a higher-dimensional Hilbert space, which enables the separation of complex patterns even with fewer samples. 
A kernel, denoted by \(k(x,x')\), serves as a measure of similarity between quantum states, when a classical input \(x\) is encoded into a quantum state. 
\[
k(x,x') = |\langle \phi(x) \mid \phi(x') \rangle|^{2}
\]
\textcolor{black}{In MC contexts, this kernel helps in the detection of subtle separations in high-dimensional sensor or telemetry data, thus distinguishing normals and anomalous satellite readings, even in the case when labeled fault samples are scarce.}

\begin{figure*}[!ht]
    \centering
    \includegraphics[width=0.6\textwidth, trim={0 0 0 1cm
    },]{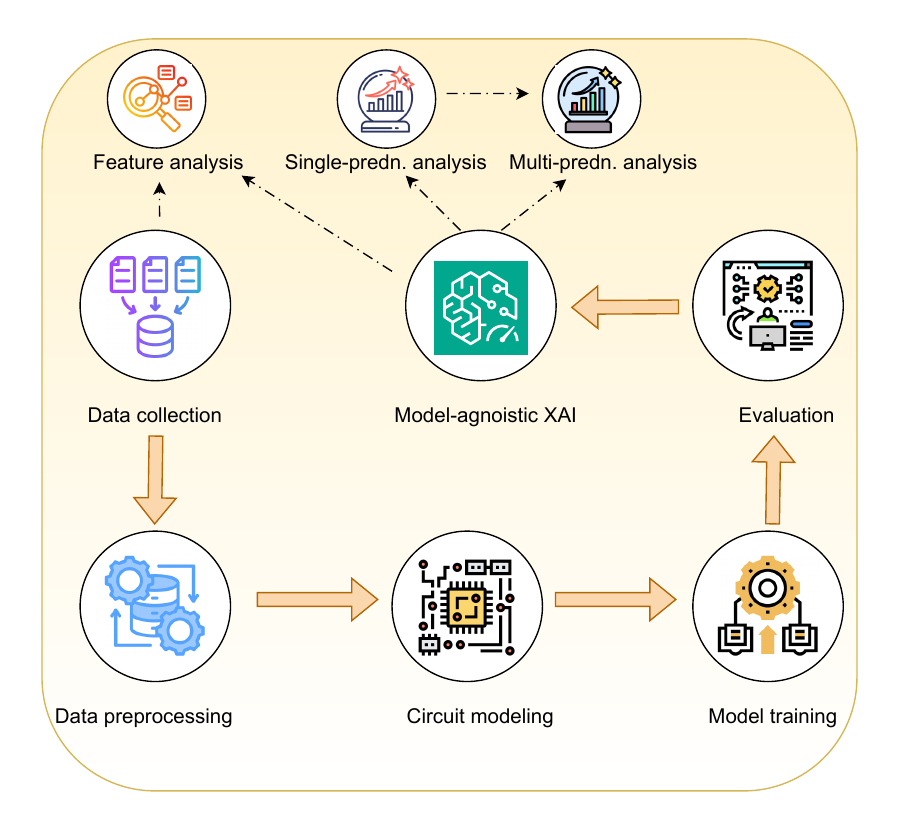}
    \caption{Workflow of a model-agnostic explainable quantum machine learning framework}
    \label{fig:qexp}
\end{figure*}

Schuld et al. \cite{schuld2019quantum} proposed that a simple encoding of the inputs into quantum states works like a non-linear feature map. The authors provide explicit constructions for such kernels. Havlíček et al. \cite{havlivcek2019supervised} implemented two quantum-classical hybrid pipelines on a superconducting processor, using a quantum variational classifier and a quantum kernel estimator combined with classical SVM. The results of this study confirm that it is possible to realize quantum-enhanced feature spaces on existing hardware to support data-scarce MC applications like integrity monitoring in safety-critical networks. Stability and predictability of training dynamics are another critical aspect in MC systems. The classical stochastic optimizers may exhibit slow convergence and instability in bad loss landscapes, which is unacceptable in real-time and embedded environments. Quantum Natural Gradient Descent (QNGD) mitigates this issue by incorporating the quantum state space geometry into parameter updates. Stokes et al. \cite{stokes2020quantum} demonstrated that QNGD allows the optimizer to move in the direction of the steepest descent with respect to the quantum information geometry. In MC applications, QNGD provides a theoretically grounded approach for stable training in long-duration missions and adaptive control in dynamic environments. 
In some of the resource-constrained MC environments, like edge devices on autonomous platforms, efficient circuit designs that meet decoherency and latency limits, and at the same time don't sacrifice model capacity. Pérez-Salinas et al. \cite{perez2020data} proposed a data re-uploading technique that satisfies the above condition. This technique embeds the same input several times within a shallow quantum circuit, which enables expressive decision boundaries with minimal depth. Data re-uploading techniques provide a low-latency solution for multiple inference tasks on satellites, autonomous vehicles, and UAVs. 

Since the current quantum hardware lacks fault tolerance, researchers have proposed hybrid quantum-classical frameworks, which are a more practical near-term option for MC deployment. Variational Quantum Algorithms(VQAs) follow this design philosophy that delegates the initial quantum state preparation and the final measurement to the quantum device and keeps optimization and loss computation on a classical processor. Cerezo et al. \cite{cerezo2021variational} provided detailed architectures, mitigation approaches, and convergence analysis of VQAs, which are compatible with noisy-intermediate scale quantum hardware. This hybrid structure of VQAs allows safety interlocks and time-critical I/O to be under classical control while asynchronously invoking quantum routines. This configuration is ideal for auditable operations in energy optimization and disaster management.

One of the notable requirements for MC integration is the credible experimental validation of learning performance on physical quantum devices. Huang et al. \cite{huang2022quantum} presented theoretical and experimental demonstrations of \enquote{quantum advantage} in learning tasks. The authors validated on 40 superconducting qubits that executed around 1300 quantum gates. They reported that the quantum systems could learn from an exponentially smaller number of experiments. These findings strengthen the fact that meaningful workloads can already be executed on real hardware, which supports calibration, diagnostics, and adaptive control in space missions and aerospace.

\subsection{Quantum Uncertainty, Robustness, and Fault Tolerance}\label{qurf}
Predictable behavior under uncertainty is a critical need of MC systems such as defense decision loops, spacecraft control, and autonomous air-traffic coordination. 
Asymptotics and heuristics are generally employed in classical systems to estimate uncertainty. These can lead to miscalibration in distribution-shifted or small-data regimes. In quantum computing, a technique called Quantum Conformal Prediction (QCP) uses conformal prediction along with the intrinsic randomness of quantum measurements and hardware noise to produce valid error limits. Park et al. \cite{park2023quantum} noted that due to device and measurement noise, their quantum models yield multiple random decisions per input. The authors leveraged randomness to define prediction sets that capture the uncertainty of the model provably, providing finite-sample coverage guarantees for both regression and classification. The authors also performed experiments on existing quantum hardware, along with simulators, to confirm the calibration guarantees. The basic building block for uncertainty is the variance of an observable parameter \(A\), measured on a state \(\rho\):
\[
(\Delta A)^2 = \langle A^2\rangle - \langle A\rangle^2, \quad \langle A\rangle=\mathrm{Tr}(\rho A)
\]
where \(Tr(.)\) denotes the trace. 

Another requirement in MC systems is the robustness to device noise during inference and training. As of current research, full-scale quantum error correction (QEC) is an asymptotic solution. Fowler et al. \cite{fowler2012surface} provided quantitative fault-tolerance estimates for surface-code logical qubits, decoding, and operations, indicating their practicality for large-scale systems. Error mitigation naturally occurs with learning/variational loops. Two popularly analyzed techniques for mitigation are probabilistic error correction and Zero Noise Extrapolation (ZNE). ZNE evaluates quantum circuit at amplified noise levels and extrapolates to zero noise \cite{temme2017error}. ZNE fit can be represented as 

\[
E(0) \approx \sum_{i=1}^{m} c_i\, E(\lambda_i), \quad \text{s.t. } \sum_i c_i=1
\]

Where \(E(\lambda_i)\) is the measured expectation at \(\lambda_i\) (amplified noise factor) and \(c_i\) represent extrapolation coefficients that are chosen to cancel low-order noise terms. Li et al. \cite{li2024ensemble} integrated the error mitigation with Variational Quantum Classifier (VQC) and demonstrated improvement in the learning. 

Resilience to adversarial effects and perturbations, combined with the tools for formal robustness verification, is another requirement in MC systems. Randomized encodings and quantum noise can certify robustness bounds of quantum classifiers (against adversarial examples) on the defensive side. Huang et al. \cite{huang2023certified} derived robustness bounds and connected them with the notion of differential privacy. Lin et al. \cite{lin2024veriqr} proposed VeriQR, which provides an automated formal verification workflow with adversarial example discovery, local/global robustness checks, and adversarial training. VeriQR can enable pre-mission certification (verification under disturbances) and in-mission monitoring that triggers fallbacks at violation of verified margins.

\subsection{Quantum Explainability and Verification}
MC deployments in disaster-response logistics, satellite operations, safety-critical transport, etc., require that the output of quantum AI be not only accurate, but also interpretable, trustworthy, and verifiable against perturbations and implementation errors. 

Formal robustness verification oriented to quantum classifiers can potentially benefit the trustworthiness of QAI decisions in MC contexts. VeriQR framework \cite{lin2024veriqr} (also discussed in Section \ref{qurf}) provides complete and sound algorithms for robustness verification (local and global) of QAI models under realistic device noise. VeriQR can automatically find adversarial examples and perform adversarial training to improve robustness. These properties provide concrete verification proofs that are comparable to classical safety cases. 
Caro et al. \cite{caro2023classical} proposed protocols for classical verification of quantum learning based on interactive-proof frameworks. These protocols allow the classical clients to verify an untrusted quantum server executing a delegated learning task in terms of correctness of execution. This establishes correctness guarantees for remote QAI. Hence, the works by Lin et al. \cite{lin2024veriqr} and Caro et al. \cite{caro2023classical} give evidence of the feasibility of delegated-execution verification and robustness checks with current QAI formulations, thus addressing the audit requirement of  MC deployments.

Recently, several research papers have been adapting feature-attribution ideas of classical ML explainability, like LIME (Local Interpretable Model-agnostic Explanations) and SHAP (Shapley Additive Explanations), to QAI models, including quantum kernels, parametrized quantum circuits, and QSVMs. Figure \ref{fig:qexp} presents the workflow of a model-agnostic explainable quantum machine learning framework. It shows how data is collected and processed to build and train a quantum circuit, which is evaluated using model-agnostic explainable AI methods. The model-agnostic explainable evaluates the trained model and feeds its interpretations to refine feature analysis and analyze both single and multi-prediction outcomes.  In this framework, Kadian et al. \cite{kadian2025exqual} proposed ExQUAL (Explainable Quantum Classifier), integrating LIME and SHAP into Pegasos QSVM to produce both global and local explanations for binary and multi-class tasks. This shows that the classical Explainable AI(XAI) methods can be combined with trained quantum models and quantum feature maps to obtain human-interpretable attributions. 
Heese et al. \cite{heese2025explaining} apply the concept of Shapley values to quantify the importance of groups of gates in a QAI model for attaining specific goals. The authors proposed that the resulting attributions can be seen as explanations for why a particular circuit works well for a given task, which further improves the understanding of how to construct both parametrized and variational quantum circuits. The authors demonstrated the benefits of the proposed framework for transpilation, classification, optimization, and generative modeling with experimental evaluation on both simulators and superconducting hardware devices. 
Kottahachchi et al. \cite{kottahachchi2025qrlaxai} proposed QRLaXAI (Quantum Representation Learning and Explainable AI), combining a quantum autoencoder with a variational quantum classifier and incorporating empirical and theoretical explainability for image data/ The authors leveraged visual explanations of LIME and analytical insights of SHAP models to attain a deeper understanding of the model's decision-making process based on output.
\textcolor{black}{
It is important to acknowledge here that, while these adaptations of classical XAI techniques to QAI models are a necessary step, they are not sufficient for MC certification.  It remains an open question whether LIME- and SHAP-derived explanations of quantum models constitute acceptable evidence under standards like IEC 61508 and DO-178C, as no certification authority has yet evaluated these quantum-specific explainability components.  Further, there is also scope for designing explainability metrics specific to quantum machine learning models \cite{pira2024interpretability}. 
}

\section{Applications} \label{appns}
\begin{figure*}[!t]
    \centering
    \includegraphics[width=0.6\linewidth]{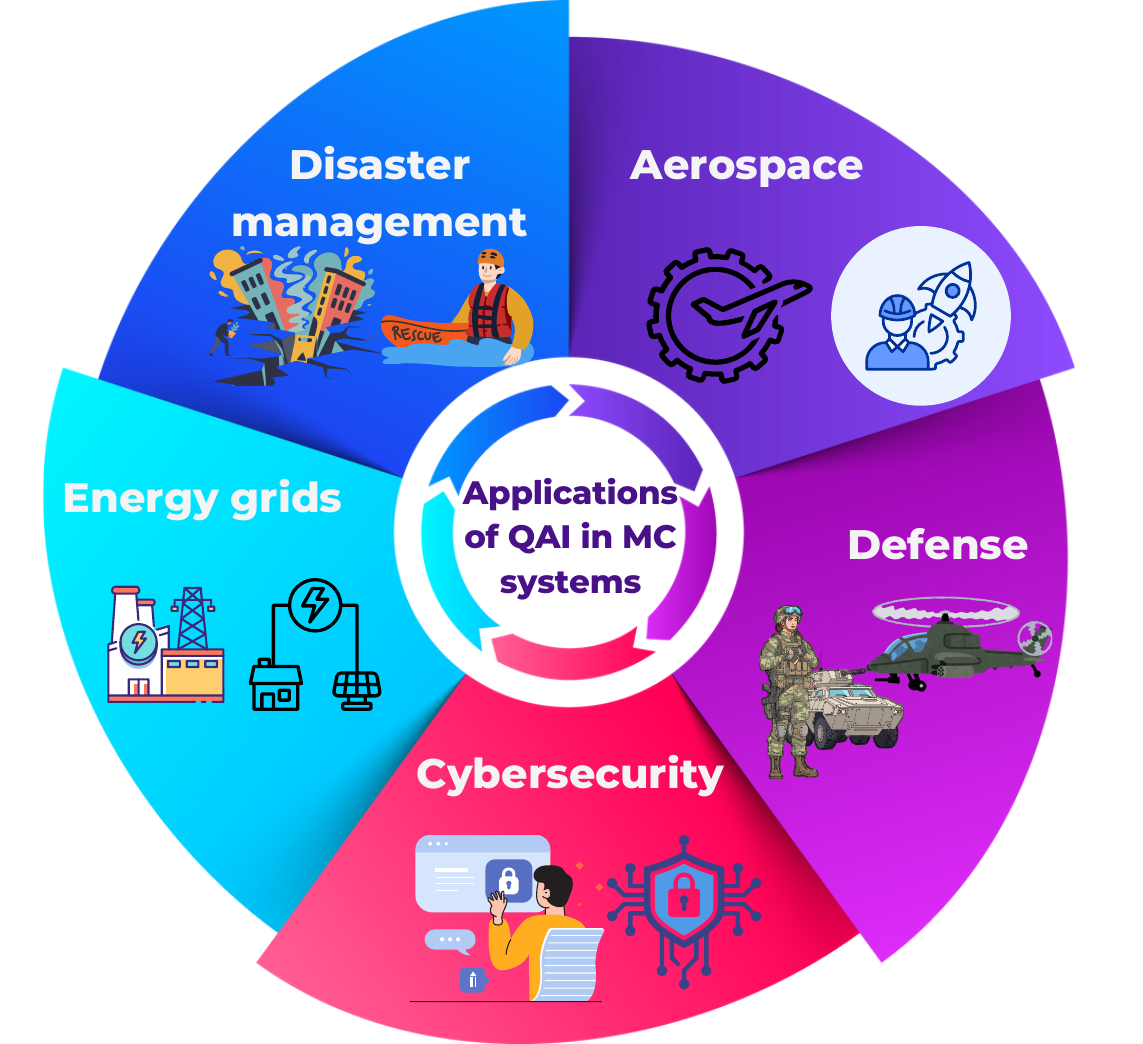}
    \caption{Applications of Quantum AI in Mission critical systems}
    \label{fig:appns}
\end{figure*}

Quantum AI has multifarious applications in mission critical systems -- some of them are highlighted in Figure \ref{fig:appns} along with their discussion below.

\subsection{Aerospace and Space Systems}
Aerospace planning tasks like trajectory design, de-confliction, and constellation scheduling are treated as large combinatorial problems. Several research works show that these tasks can be mapped to quadratic unconstrained binary optimization (QUBO) and solved by exploiting quantum heuristics, such as variational approaches and annealing. 
\textcolor{black}{
QUBO is a combinatorial optimization formulation where binary decision variables are optimized over a quadratic objective without any explicit constraints. In QUBO, all problem constraints are absorbed into the objective as penalty terms. QUBO is natively compatible with variational quantum algorithms such as QAOA and quantum annealing hardware.}
Stollenwerk et al. \cite{stollenwerk2019quantum} constructed a conflict-graph formulation for strategic air traffic de-confliction. The authors showed how wind-optimal multiple flights can be mapped to QUBO and further evaluated the problem instances on classical baselines and D-Wave hardware, thus demonstrating an end-to-end encoding and solution workflow. For spacecraft trajectory optimization, Grossi et al. \cite{de2025transcription} transcribed continuous-time dynamics into a binary QUBO and proposed to solve interplanetary transfers using quantum annealing. The authors also dealt with discretization and constraint-handling steps that are required to obtain feasible trajectories. Makhanov et al. \cite{makhanov2024quantum} proposed a modular hybrid model for flight-path optimization and analyzed resource constraints in quantum backends and classical simulators, clearly indicating where the quantum subroutines can be fitted in present-day pipelines. In all these formulations, the standard QUBO takes the form: 
\[
E(\mathbf{s}) = \mathbf{s}^{\mathsf T} Q \mathbf{s}
\]
 where \(s \in {0,1}^n \) are decision variables (such as discrete heading, timing, assignment choices) and \(Q \in \mathbb{R}^{n \times n}\)
 encode penalties and costs derived from hard constraints and the objective of the mission, such as slew limits, fuel and separation minima.

Aerospace operations demand quantified uncertainty and reliable fault detection with telemetry systems. Quantum anomaly detection methods provide procedures to detect variations from the normal behavior after encoding the classical data into quantum feature spaces. Liang et al. \cite{liang2019quantum} proposed a quantum anomaly detection method based on density estimation and multivariate Gaussian distribution. The authors constructed both algorithms using the standard gate-based model of quantum computing. They also presented a quantum procedure for efficiently estimating the determinant of any Hermitian operator. 

Multi-agent decision-making under partial information is central in autonomous spacecraft control and coordinated swarms. Chen et al. \cite{chen2024qmarl} introduced a quantum multi-agent reinforcement learning framework (QMARL), which functions based on quantized state/action representations and reports coordination and navigation experiments in multi-robot settings. The authors provided implementation details of the training loop for decentralized agents and PQC-based policy modules. Although the aerospace swarm-flight optimization is an engineering task, the study of Chen et al. provides a concrete algorithmic template (state encoding, policy circuit design, and training protocol) for migrating multi-agent navigation/avoidance to quantum-enhanced controllers. Further, the classical literature \cite{sun2023distributed} on distributed cooperative control for spacecraft swarms provides the connectivity models and the constraints for quantum reinforcement learning (QRL) models to follow for collision-avoidance and connectivity maintenance. This literature also clarifies the interface between the learning-based policies and the guidance laws. 

\subsection{Defense and Autonomous Systems}
Defense autonomy combines multi-sensor perception, mission planning, and adaptive control under strict timing and safety constraints. The planning layer in these autonomous systems is generally reduced to NP-hard scheduling and routing, and the autonomy part balances fuel/time budgets, online re-planning under uncertainty, and collision avoidance \cite{evers2014online}. 

Under quantum optimization for mission routing and tasking, Mori et al. \cite{mori2023quantum} formulated a real routing problem as a QUBO and solved it using quantum clustering and annealing-based pre-processing. The paper presents an end-to-end encoding and reports the solution quality on practical instances, thus illustrating how logistics-type planning in defense autonomy can be mapped onto annealing hardware. The authors demonstrate that the annealing-based solvers can easily handle real constraints like capacities and time windows within hybrid pipelines. For sensor-driven navigation and collision avoidance, Sinha et al. \cite{sinha2025nav} developed quantum deep reinforcement learning agents that included a policy based on PQCs. The authors evaluated the proposed framework on autonomous navigation tasks and reported improved training stability(less sensitivity to initialization of the weights), a higher average cumulative reward compared to classical baselines, and explicit deterioration under injected quantum noise, thus quantifying the robustness tradeoffs. In the proposed framework, the optimized target is the discounted return:
\[
J(\theta) = \mathbb{E}\!\left[\sum_{t=0}^{T}\gamma^{t} r_t \,\middle|\, \pi_\theta \right]
\]

where \(r_t\) is the reward at time t, \(\gamma\) is the discount factor, \(\pi_\theta\) is the PQC-parametrized policy with parameters \(\theta\), and \(T\) is the episode horizon. In defense autonomy, a requirement is that the policies must retain performance under real-world disturbances and hardware noise; hence, the work of Sinha et al. is directly relevant because of its noisy-simulation study.

For model reliability under device noise, Heyraud et al. \cite{heyraud2022noisy} characterized how dephasing and dissipation impact quantum-kernel classifiers and support vector machines. The authors quantified performance loss as a function of circuit depth and noise parameters, and discussed regularization and mitigation strategies. The results of this work provide guidance on operating regimes (where kernel-based QAI remains reliable) for defense sensor-fusion stacks adopting quantum kernels. 

\textcolor{black}{
The upcoming Agentic AI (AAI) technology can potentially benefit the multi-UAV defense applications in several ways. Khowaja et al. \cite{khowaja2025integration} demonstrated that AAI systems can be used for autonomously analyzing multimodal sensor data, dynamically allocating resources, and coordinating responses in real time, thus reducing initial response times by over 60\% compared to rule-based systems. The integration of QAI techniques can potentially accelerate the capabilities of AAI-based UAV systems by enabling exponentially faster optimization of resource allocation and threat prediction across vast and complex operational environments that might overwhelm classical computing approaches.}

\subsection{Cybersecurity and Critical Infrastructure Protection}
Critical infrastructure networks like power, water, and transportation require timely detection of advanced threats and cryptographic resilience in an increasing manner. Quantum Key Distribution (QKD) can be used in utility fiber networks for high-assurance keying; further, quantum keys can be used for authentication of grid communications. \textcolor{black}{QKD is a cryptographic protocol that exploits quantum mechanical properties like the no-cloning theorem and measurement disturbance in order to generate a shared secret key. The security of this key is guaranteed by the laws of physics rather than computational hardness assumptions.} Intrusion Detection Systems (IDS) in cybersecurity are increasingly relying on machine learning; the quantum machine learning tools can potentially replace them in areas or situations where a significant quantum advantage might be reaped \cite{sai2025quantumcs}. Secure collaboration across organizations demands privacy-preserving learning, and the recent quantum homomorphic encryption (QHE) based decentralized learning frameworks provide data privacy guarantees that are computation-theoretic.

Alshowkan et al. \cite{alshowkan2022authentication} made use of QKD protocols in the authentication of smart grid communications, which are executed on a deployed electric-utility fiber network. The authors created a software stack that ingests keys from a QKD system and uses them to generate message-authentication codes for control links in machine-to-machine scenarios. Thus, the proposed framework strengthens operational traffic with keys that have physical-layer security properties. Kumar et al. \cite{kumar2025quids} proposed QuIDS, an intrusion detection system based on a quantum support vector machine. The authors proposed to detect IoT attack classes using small training sets of low-dimensional features. The authors demonstrate competitive results against benchmarks in IoT intrusion detection systems. The decision function of QSVM follows the standard large-margin form: 
\[
f(x) = \operatorname{sign}\!\left(\sum_{i=1}^{m}\alpha_i\, y_i\, k(x_i,x) + b\right)
\]

where \(x\) is the feature vector, \({(x_i,y_i)}\) for \(i=1\) to \(m\) are the support vectors \(x_i\) with labels \(y_i\), \(k(.,.)\) is the quantum kernel, \(b\) is the bias, and \(\alpha_i\) are the learned dual weights. 
\textcolor{black}{In MC cybersecurity settings, this decision function helps in classifying the incoming network flows as benign or malicious. The quantum kernel can improve the  separation of attack signatures from classical feature spaces.} 

Kalinin et al. \cite{kalinin2023security} also evaluated QSVMs and quantum convolutional neural networks (QCNNs) against classical IDS systems on standard traffic datasets. The authors encode the flow/packet features into PQCs and compare performance under matched training budgets. The paper not only shows the construction of quantum feature maps for network security data, but it also quantifies cases like limited data, where quantum kernels are competitive with classical detections. 

Li et al. \cite{li2025quantum} proposed a QHE based federated learning framework for cross-operator collaboration without the need to expose raw data. In this scheme, the clients encrypt quantum data and delegate tracing to a server, and get outputs that have computation-theoretic privacy \cite{sai2025quantumfl}. The authors demonstrated that the proposed framework reduces communication complexity significantly compared to blind quantum computing approaches. 

\begin{figure*}[!ht]
    \centering
    \includegraphics[width=0.8\textwidth, height=3.5in]{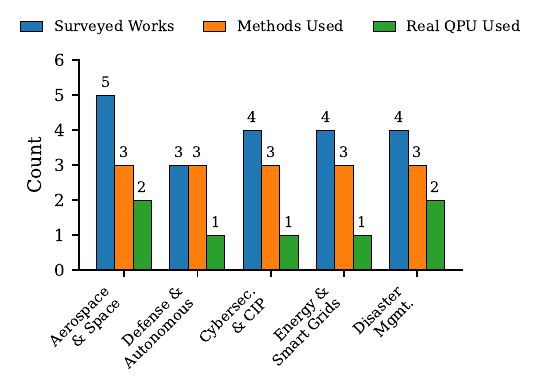}
    \caption{\textcolor{black}{Comparative Survey Summary Across MC Application Domains}}
    \label{fig:analysis}
\end{figure*}

\subsection{Energy and Smart Grids}
The operation of smart grids require reliable and fast decisions for stability and situational awareness. The classical frameworks are stressed mainly due to the fact that the core optimizations are computationally hard (AC optimal power flow is strongly NP-hard), as shown in research \cite{bienstock2019strong}. This complicates real-time convergence and guarantees of classical solvers. The variation in solar and wind energies provides steep load ramps that require multi-interval flexibility from conventional resources \cite{caiso}. 
Recent research explored quantum reinforcement learning to assist stability control and contingency analysis in smart grids. Peter et al. \cite{peter2025quantum} applied QRL for power grid security assessment, particularly for combinatorially difficult contingency analysis problems (where classical RL fails in terms of computational cost). The proposed QRL framework helped scale by improving computational efficiency and agent proficiency. The authors demonstrated a proof-of-concept by using QC for RL agent simulation and training. The proposed hybrid agent runs on quantum hardware based on IBM's Qiskit runtime. The optimized target in the proposed framework is the discounted return.

\begin{table*}[!ht]
\begin{tabular}{|p{2.5cm}|p{4cm}|p{6.5cm}|p{2cm}|}
\hline
\textbf{Companies involved} & \textbf{Topic of collaboration} & \textbf{Brief description on the work} & \textcolor{black}{\textbf{Stage}} \\
\hline
SandboxAQ \& United States Air Force (USAF) & AI-assisted magnetic navigation (Quantum) & Testing AI-assisted navigation using quantum magnetometers as a primary guidance method (without GPS) & \textcolor{black}{Field demo} \\
\hline
Air Force Research Laboratory (AFRL) \& QC Ware & Quantum Machine Learning (Q-means) & Exploring quantum clustering (q-means) to group unmanned aircraft flight paths and infer intent. & \textcolor{black}{PoC} \\
\hline
Airbus, QC Ware \& IonQ & Aerospace optimization (Hybrid Quantum) & Running hybrid quantum programs for cargo loading, routing, and other optimization problems. & \textcolor{black}{PoC} \\
\hline
IBM \& Raytheon & Quantum computing and AI for defense & Collaborating to co-develop solutions that integrate QC and AI particularly for applications in the defense sector. & \textcolor{black}{Exploration} \\
\hline
Terra Quantum \& (Utility partners) & Energy time-series forecasting (Hybrid Quantum) & Developing hybrid quantum models to improve short-term predictions for energy metrics for grid balancing. & \textcolor{black}{PoC} \\
\hline
D-Wave, TNO \& Quantum Quants & Grid optimization (Hybrid Quantum) & Developing a hybrid quantum constrained quadratic model solver for energy grid redistribution and partitioning. & \textcolor{black}{PoC} \\
\hline
IBM \& E.ON & Utility operations and forecasting (Quantum/QML) & Exploring quantum methods for complex utility operations. & \textcolor{black}{Exploration} \\
\hline
IonQ \& Hyundai & 1. Battery chemistry (Quantum) \newline 2. Machine perception (Quantum ML) & 1. Using variational quantum eigensolver to study lithium compounds. \newline 2. Experimenting with QML for machine perception. & \textcolor{black}{PoC} \\
\hline
\end{tabular}
\caption{A summary of industrial practice of \textcolor{black}{QAI} in MC systems \textcolor{black}{(PoC: proof-of-concept without production deployment; Field demo: tested in an operational environment; Exploration: early-stage collaboration without published results)}}
\label{summaryindustry}
\end{table*}

Unit commitment (UC) and Optimal Power Flow (OPF) remain critical for scheduling and resource allocation in smart grids. \textcolor{black}{Unit commitment is a short-term scheduling optimization problem that decides which of the generating units should be turned off or on over a planning horizon, which typically ranges from 24-168 hours, with the aim of meeting the forecast demand at minimum cost. It turns out to be a large-scale mixed-integer optimization problem after incorporating operational constraints like minimum up/down times, ramp rates, and startup/shutdown costs.} \textcolor{black}{Optimal Power Flow is a steady-state network optimization problem that decides the optimal operating point of a power system with minimization of an objective (generally, generation cost or transmission losses), while satisfying both physical laws (Kirchhoff's laws) and operation constraints (line thermal limits, voltage limits).}

Salgado et al. \cite{salgado2024hybrid} developed a hybrid quantum-classical algorithm to approximately (efficiently) solve the UC problem in polynomial time (the UC problem is NP-hard). The authors decomposed the UC problem into two subproblems - a QUBO and a quadratic optimization problem. The authors made use of QAOA to determine the optimal unit combination, along with classical methods to identify individual unit powers. The proposed framework reduced the number of iterations required for QAOA convergence and also improved the solution accuracy. 

For fault prediction and anomaly detection, Ajagekar et al. \cite{ajagekar2021quantum} proposed a QC-based deep learning framework. The deep learning stack consists of a conditional Boltzmann machine for feature extraction and deep neural networks for classification. The authors addressed the computational challenges of training deep learning models with QC-based training methodologies. \textcolor{black}{The authors reported improved computational efficiency and faster response time with the proposed hybrid model on their specific benchmark, though these gains remain to be validated at production scale.}

\subsection{Disaster Management}
Disaster management requires logistic planning, early warning, infrastructure restoration, and rapid evacuation under severe uncertainty. The classical formulations with evacuation scheduling and routing, emergency vehicle routing, and post-disaster restoration are treated as NP-hard, which leads to heuristic screening or approximations in time-critical settings \cite{islam2022simulation}. In this landscape, \textcolor{black}{QAI} can potentially provide significant advantages with quantum-accelerated restoration scheduling, quantum optimization for logistics/evacuation, and quantum-enhanced early warning systems. 

Liu et al. \cite{liu2023performance} conducted a performance evaluation of tsunami evacuation route planning by solving a combinatorial optimization problem on quantum and non-quantum annealers. The authors modeled the evacuation problem mathematically and converted it into QUBO form. The authors evaluated the performance of four annealing machines, including D-Wave Neal, D-Wave Advantage, Fixstars Amplify Annealing Engine, and Vector Annealing. They conducted an evaluation from three aspects: the solution quality, time to solution (TTS), and the maximum number of evacuees. Fu et al. \cite{fu2023coordinated} proposed a hybrid quantum-classical approach for coordinated post-disaster restoration (CPR). The authors established the CPR model with the consideration of power distribution system (PDS) operation, topology configuration, microgrid operation, and repair crew dispatch. They made use of binary expression and integer slack methods to transform the constrained problem into QUBO. D-Wave's direct quantum processing unit solvers and hybrid solvers of Leap quantum cloud service are employed to perform a comparative analysis of different calculation methods. Sankaradass et al. \cite{sankaradass2025leveraging} proposed QSVM and QNN-based models for early warning systems for environmental disaster prediction. The authors tested their proposed models on two benchmark datasets, and the findings reveal that the proposed QAI models are accurate than the classical ML models. Kobayashi et al. \cite{kobayashi2020r} outlined an evacuation planning stack that combined the D-Wave quantum annealing engine with the NEC SX-Aurora vector supercomputer. They presented a high-level workflow of evacuation models, which consists of the initial QUBO mapping and partitioning of computation between the annealer for the last-stage optimization and classical vector hardware. They also provide a concrete systems blueprint for integrating HPC evacuation simulators with annealing-based optimizers. 

Figure \ref{fig:analysis} presents a bar chart quantifying the surveyed works, methods used, and real QPU used across five MC domains. The surveyed works on aerospace used three methods - QUBO/annealing, anomaly detection, QMARL; works on defense used QUBO/annealing, QRL, quantum kernels; works on energy used QRL, QAOA, quantum boltzmann machines; and works on disaster management used QUBO/annealing, QSVM, and QNN.

\section{Industrial Practice}\label{industry}
 Several companies are actively exploring and developing proofs-of-concept (PoCs) on application of QAI methods to solve challenges in MC systems, particularly in aerospace and defense. They are using simulated quantum computers and current-day NISQ quantum computers. A summary of industrial efforts in this area is noted in Table \ref{summaryindustry}.

SandboxAQ \cite{sandboxaq_2025}, in partnership with the United States Air Force (USAF), is testing AI-assisted magnetic navigation that uses quantum magnetometers to navigate, without the need for FPS. The USAF flew the system from Joint Base Charleston, where magnetic navigation was used as the primary guidance method during the flight, thus reporting a real-time demo. Air Force Research Laboratory (AFRL) partnered with QC Ware to explore quantum machine learning clustering, termed q-means, to group unmanned aircraft flight paths and infer intent. This is an early research towards situational awareness and autonomy. Airbus, in partnership with QC Ware and IonQ, runs quantum programs on cargo loading, routing, and other related aerospace optimization problems \cite{airbus_2024}. The aim of this project is to examine how hybrid quantum approaches could surpass or complement existing High Performance Computing (HPC) heuristics. IBM and Raytheon (an Air Force and defense tech company) forged a collaboration to co-develop solutions that combine quantum computing and AI for the defense sector \cite{ibm}. Terra Quantum is developing hybrid quantum models for energy time series forecasting, such as steam flow from power plants and solar output \cite{terra}. The aim of the project is to improve the short-term predictions for grid balancing and operations planning with utility partners. D-Wave is partnering with TNO and Quantum Quants to develop hybrid quantum constrained optimization on redistribution and grid partitioning scenarios. Their hybrid constrained quadratic model solver reported to find high-quality solutions within fixed time budgets, compared to several classical baselines \cite{dwave}. IBM is collaborating with E.ON (Europe's largest energy company) to explore quantum methods for complex utility operations and forecasting at scale, plausibly using QAI techniques. The main focus is planning and optimization under renewable energy variability, which is being dealt with as exploration rather than a production rollout. IonQ and Hyundai are working on two separate research threads relevant to future transportation systems \cite{ionq_hyundai_2024}. Firstly, they are planning to develop next-generation battery chemistry technology by using Variational Quantum Eigensolver (VQE) to study lithium compounds, which also aims at improving the safety and performance. The companies are also pursuing quantum ML experiments for machine perception, starting with road-sign classification, and moving towards 3D point-cloud object detection.

\section{A Model for Quantum Resource Management and Scheduling System}\label{framework}
There has been extensive research in the classical/high-performance computing cloud scheduling and resource management, but quantum cloud computing, being nascent, lacks in this. It is not possible to completely use the classical scheduling and resource management systems in quantum cloud computing, due to the different nature of the quantum tasks. Hence, we propose a framework to handle the unique properties and constraints of quantum computing in resource management and scheduling systems.

\textcolor{black}{We assume the following deployment conditions for the framework. (i) a heterogeneous pool of NISQ-scale QPUs (up to a few hundred qubits) are accessible via cloud APIs, (ii) data regarding real-time calibration like gate fidelities and coherence times is available from each QPU, (iii) the quantum jobs cannot be checkpointed or preempted in between execution, and (iv) classical pre-processing, post-processing, and orchestration are co-located with the scheduling service.}

\begin{figure*}[!ht]
    \centering
    \includegraphics[width=\textwidth, trim={0 0 0 0
    }]{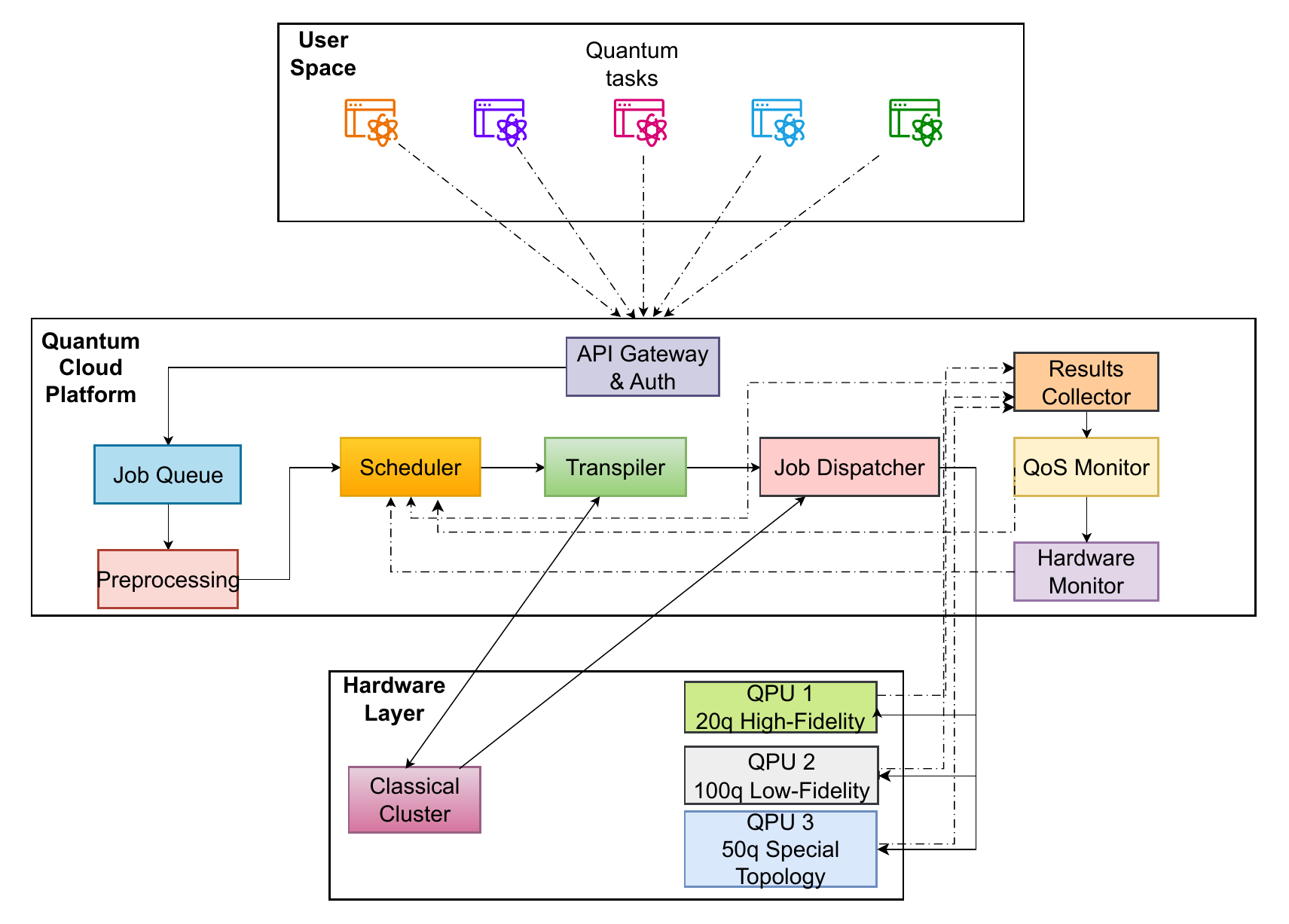}
    \caption{Proposed Quantum Cloud Scheduling and Resource Management Framework}
    \label{fig:arch}
\end{figure*}

Figure \ref{fig:arch} shows the proposed software framework. It begins with \texttt{API Gateway and Authorization} component, which acts as a secure front door for all the interactions of users. Apart from the standard authentication and quota management, this component is particularly proposed to ingest complex quantum task definitions, including serialized objects from frameworks such as Cirq or Qiskit and the circuits in QASM format. This component also accepts other user-defined preferences and constraints, including the minimum acceptable 2-qubit gate fidelity, required qubit count, the maximum acceptable queue wait time, and the priority of the task. All of these metadata are essential for downstream decision making.

After validation, a job will be passed to a \texttt{Job Queue}, which is a scalable and persistent buffer that separates the intensive scheduling process and high-volume submission of tasks. Although this job queue is functionally similar to classical queues, it will be optimized to store and prioritize jobs based on their quantum metadata. The \texttt{Preprocessing \& Analysis service} retrieves jobs from the job queue to execute a \enquote{quantum-aware} inspection. This component estimates several fundamental properties of the user's circuit by performing static analysis, including the circuit depth, the exact qubit count, and the number of high-error-rate 2-qubit gates. An important function of this component is to generate an accurate estimate of the quantum execution time and compare it against known coherence times, which helps in flagging jobs that are too deep to be successful. Based on this analysis, it assigns tags to jobs (like type-deep circuit, require-high fidelity) that are very helpful to the scheduler. 

Like in classical scheduling, the system requires a real-time view of the hardware, which needs to be supported by the \texttt{QPU Hardware Monitor}. Compared to a classical CPU monitor, this component is very complex and it continuously polls all the QPUs to track both the static properties like qubit count and topology, and the other dynamic calibration-dependent properties. This monitor also produces data such as 1-qubit and 2-qubit gate fidelities and current gate execution times. The monitor also reports the QPU's state (online, offline, or recalibrating), making sure that the scheduler never uses inaccurate or stale data. The \texttt{Scheduler} component is the most important component that takes in the data from the hardware monitor and decides the jobs to run on different QPUs. It also takes into special consideration quantum jobs. For example, the quantum jobs cannot be paused and resumed later with the help of some kind of checkpointing mechanism. Firstly, a feasibility filter narrows the QPU choices by finding all those that are available and that meet the job's requirements. Next, a policy engine will be used to rank the feasible QPUs to select the best. In this work, we adapt a classical HPC strategy - shortest job first, which uses the estimated execution time to maximize throughput. Other classical policies like round robin, shortest remaining time first, and multilevel feedback queues could also be adopted based on the requirement. The scheduler also considers quantum-aware policies like best-fit fidelity that try to match a job to a QPU just above the required fidelity. This helps in saving the high-performing and rare QPUs for the jobs that require them the most.
\textcolor{black}{For $n$ queued jobs and $k$ available QPUs, the feasibility filter evaluates each job against each QPU, resulting in $O(nk)$ complexity. The subsequent policy ranking sorts feasible candidates in $O(n \log n)$ for priority-queue-based policies like shortest job first.}

The jobs will be sent to the \texttt{Transpiler service} after the scheduler makes its selection, which performs heavy classical computation to translate the user's logical circuit into a physical circuit for the specific QPUs. Since there are different types of quantum hardware, the transpiler should be able to translate the user's code to the required QPU code. It is possible that the conversion process adds error and depth, so a sophisticated scheduler may use a trial transpilation to select the QPU that gives the lowest swap overhead. After this, the final, machine-ready program will be handed over to the \texttt{Job Dispatcher}, which opens a connection to a QPU and executes the program for a particular number of shots. 
The raw results will be collected by the \texttt{Results Collector \& Post-processing} service, whose important quantum-specific role is to apply error mitigation techniques, analogous to the classical post-processing steps that use calibration data to correct for known measurement errors, before handing the final answer to the user. A \texttt{QoS Monitor and Feedback} loop observes the entire system's performance, tracking metrics like the job wait time that help the processing service improve its execution time estimates for future jobs, and also the hardware monitor by indicating any QPU that might be performing poorly.

\textcolor{black}{ Table \ref{tab:classicalvsquantum} compares the classical cloud/HPC schedulers and the proposed quantum scheduling framework, that highlight the key differences that necessitate a quantum-aware design.}
\begin{table*}[!ht]
\centering
\textcolor{black}{
\begin{tabular}{|p{3.5cm}|p{5.5cm}|p{5.5cm}|}
\hline
\textbf{Aspect} & \textbf{Classical Cloud/ HPC Schedulers (e.g., SLURM, Kubernetes)} & \textbf{Proposed Quantum Framework} \\
\hline
Hardware awareness & Static specifications (CPU cores, memory, GPU type) & Dynamic calibration data (gate fidelities, coherence times, qubit topology) \\
\hline
Job preemption & Supported via checkpointing and migration & Not possible since quantum state collapses upon measurement \\
\hline
Scheduling inputs & CPU/memory/GPU requirements and priority & Qubit count, circuit depth, minimum fidelity, coherence budget \\
\hline
Error handling & Retry, checkpoint-restart, migration to a healthy node & Full re-execution required since quantum states can't be checkpointed and error mitigation is applied in post-processing \\
\hline
Compilation/Transpilation & Not required or architecture-independent & Required; circuit must be transpiled to target QPU's native gate set and topology \\
\hline
Resource heterogeneity & Largely homogeneous clusters or labeled node pools & Fundamentally heterogeneous (different qubit technologies, topologies, and noise profiles) \\
\hline
\end{tabular}
}
\caption{\textcolor{black}{Qualitative comparison of classical cloud/HPC schedulers and the proposed quantum scheduling framework}}
\label{tab:classicalvsquantum}
\end{table*}
\textcolor{black}{
The proposed framework should be capable of handling several failure modes that are unique to quantum systems. Since quantum states cannot be checkpointed, a job on a QPU that went offline must be executed on another QPU. Stale calibration data might cause the scheduler to assign a job to a degraded QPU, thus resulting in output below the required fidelity. The QoS feedback loop tries to mitigate this by triggering recalibration queries with the drop in result quality. The QPU's coherence time may be exceeded by a transpiled circuit, thus silently producing noisy output. The preprocessing service guards against this by comparing estimated execution time to coherence limits before scheduling. Under a heavy load, high-fidelity QPUs can become bottlenecks. By routing jobs that do not require top-tier fidelity to lower-performing but available QPUs, the best-fit fidelity policy alleviates this.
}
Several considerations that are applied in classical scheduling and resource management cannot be applied in a quantum setup, as of today's position of quantum research. For example, allocating jobs based on the speed-accuracy tradeoff required by a user is a common practice in classical systems. However,   Nakajima et al. \cite{nakajima2024speed} state that the existence of a speed-accuracy tradeoff in quantum systems is still doubtful. Scheduling together the tasks of different intensiveness (data-, memory-, and compute-intensive) is also a common practice to achieve higher efficiency in classical systems. However, the state of research towards quantum memory itself is still nascent \cite{naik1705random}. As the research in quantum progresses, all these aspects could be considered in quantum scheduling systems as well. A recent study \cite{carrera2024combining} showed that it is possible to use multiple quantum computers as one, thus enabling the division of a quantum task among multiple quantum systems, which is a key direction in terms of scheduling and resource management in quantum systems.

\textcolor{black}{
\textbf{Limitations of the proposed framework.} We explicitly acknowledge that the framework presented in this section is conceptual and architectural. The framework has not been implemented, prototyped, or empirically validated on any quantum cloud platform. The reported complexity bounds are analytical estimates, not measured runtimes. We have discussed the failure-modes in a qualitative fashion. The real-world performance, including end-to-end scheduling latency, the accuracy of fidelity-aware allocation, and recovery behavior under QPU recalibration or outage events, will depend on several factors like workload mix, QPU heterogeneity, and the maturity of cloud APIs exposed by quantum providers. Therefore, we present the framework as a structured starting point for the community rather than a validated production system. We also identify empirical evaluation on emerging quantum cloud platforms (such as IBM Quantum, AWS Braket, and Azure Quantum) as a future work.}

\section{Challenges and Future Directions} \label{chall}
\begin{figure*}[!h]
    \centering
    \includegraphics[width=0.5\textwidth, trim={0 0 0 0
    }]{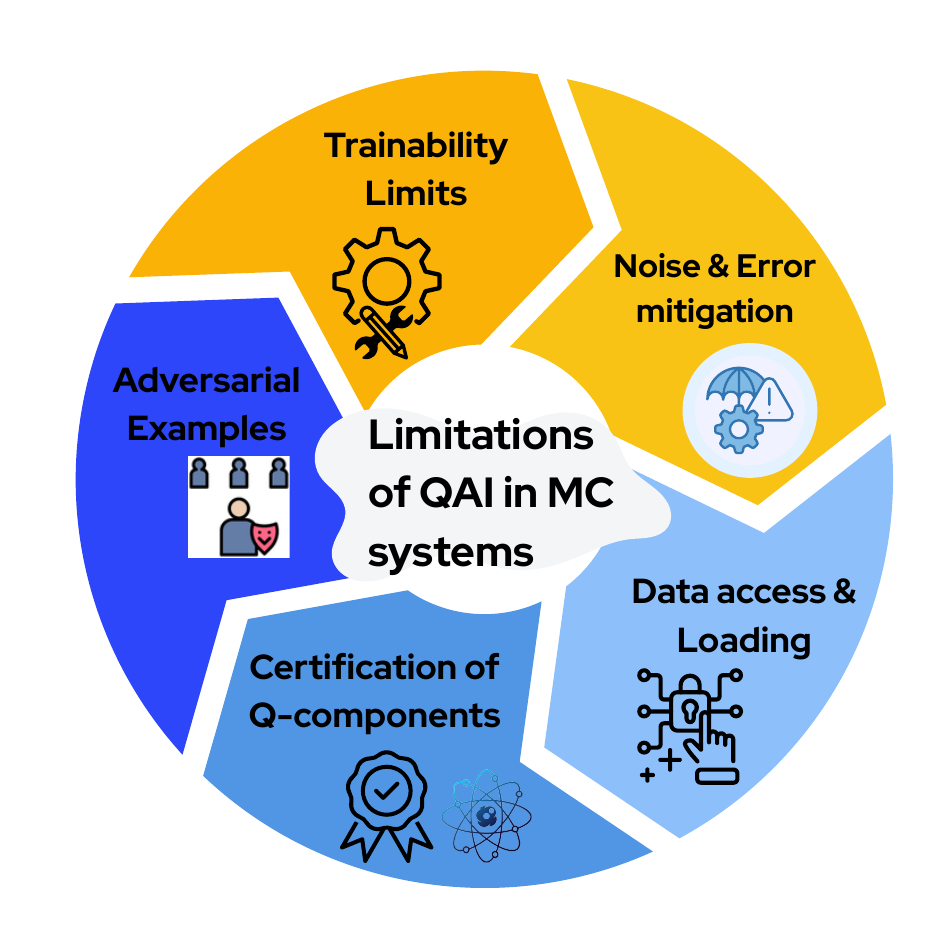}
    \caption{Limitations of Quantum AI in mission critical systems}
    \label{fig:limits}
\end{figure*}

Despite the several benefits of QAI in MC systems, the field is plagued with multiple challenges that future research should focus on. Some critical limitations of QAI in MC systems are shown in Figure \ref{fig:limits} and discussed below along with future directions.

\textcolor{black}{A fundamental tension exists between the capabilities of QAI and MC requirements. MC systems demand determinism, reproducibility, and certifiability, the properties that are challenging for noisy and probabilistic NISQ devices to provide. Quantum measurements are inherently stochastic and device noise introduces run-to-run variability on current hardware. Consequently, near-term QAI is best suited for non-safety-critical subtasks and subsystems within MC pipelines. A few examples of such tasks are offline optimization, training-time feature extraction, pre-mission planning. As of current state of QAI research, safety-critical real-time decisions should remain under deterministic classical control. Full integration of QAI into safety-critical decision loops awaits fault-tolerant quantum hardware capable of providing the reproducibility guarantees that MC certification demands.}

\noindent \textbf{Trainability limits - barren plateus:}
Variational quantum algorithms suffer from the barren plateaus problem - exponentially vanishing gradients with the increase in system size. This impedes hyperparameter tuning and learning. For broad classes of randomized PQCs, the expected gradient tends to concentrate near zero as qubit count increases, which makes optimization intractable without any tailored initialization or structural bias. Recent research \cite{wang2021noise} also shows that noise induces barren plateaus, thus placing a fundamental limit on trainability on NISQ-based hardware. The MC pipelines that rely on PQC-based training, like adaptive control or classification, must adopt provably barren plateaus, avoiding designs and shallow ansätze (problem-informed) to maintain learnability.

\noindent \textbf{Noise and error mitigation: }
Error mitigation techniques like zero-noise extrapolation and probabilistic error cancellation are able to reduce the bias without the need for full error correction. However, they incur runtime and sampling overheads that grow with circuit size and noise strength. Long-term reliability of MC systems requires fault-tolerant quantum computing. Hence, techniques like surface codes that quantify thresholds and the physical-qubit overheads to realize logical gates and qubits should be analyzed in MC systems. Error mitigation may just stabilize short-depth workloads in MC systems. 

\noindent \textbf{Data access and loading bottlenecks: }
There is a presumption of access to large classical datasets in quantum superposition in many stated QAI advantages. Quantum random-access memory (QRAM) architecture describes how \(2^n\) cells can be accessed with \(n\) address qubits with only \(O(log N)\) active switches per query. However, this is a very non-trivial hardware assumption that highlights the data-loading problems as a distinct systems challenge \cite{giovannetti2008quantum}. Therefore, the MC systems with significant classical telemetry (like sensor fusion) must favor QAI formulations or engineer efficient state-preparation pipelines with less data-loading demands. 

\noindent \textbf{Certification and verification of quantum components: }
Before the quantum model can be deployed in the real-world mission critical systems, it is important to check that the model running in production is the same as validated in the lab. Hence, research must be directed towards developing tools for circuit equivalence checking that perform an automated check and prove if two circuits implement the same transformation after their compilation. QCEC (Quantum Circuit Equivalance Checking) tool \cite{burgholzer2021qcec} provides a good starting point in this direction. On the other hand, model-level verification frameworks need to be developed, which enable testing the robustness of a quantum classifier with a particular given input. Lin et al. \cite{lin2024veriqr} provided VeriQR in this direction. 

\noindent \textbf{Adversarial examples and robustness of QAI models: }
Recent research revealed that the quantum machine learning models can be fooled by tiny and carefully crafted input changes (adversarial examples), analogous to the classical machine learning models. Deploying such weak QAI models in MC systems can have severe consequences. Although the research has progressed in the lines of quantification of robustness of QAI models \cite{gong2022universal}, it is far beyond accurate adversarial robustness. The QAI models must therefore be treated as regular security-critical software, with routine stress and adversarial attack scenario tests. This is required until foolproof QAI models are developed.

\textcolor{black}{
The uncertainty of MC systems can be potentially handled by an alternative paradigm- fuzzy logic, which processes graded truth values through if-then rules that are human interpretable. Fuzzy logic offers a potential advantage for MC deployment in terms of its interpretability. Gentil et al. \cite{gentili2025exploring} proposed chemical pathways for fuzzy quantum computing. However, fuzzy systems require design of membership functions and rule bases by experts, they scale very poorly to high-dimensional input spaces, and they don't have the ability to learn representations from raw data. At the cost of stochasticity, limited interpretability, and certification challenges on current hardware, QAI can automatically extract features from high-dimensional sensor data and discover complex non-linear patterns. Thus, the two approaches are potentially complementary in nature, with fuzzy logic being better suited for subsystems with well-codified domain knowledge, while QAI might offer advantages for data-rich subtasks like anomaly detection and pattern recognition, which have already seen the deployment of classical ML. }

\section{Conclusions}
Quantum Artificial Intelligence has a significant potential to redefine the operational foundations of mission critical systems by combining the intrinsic parallelism and non-classical computation of quantum computing with the probabilistic reasoning of AI.  In this paper, \textcolor{black}{we examined how QAI offers pathways to potentially address specific uncertainty management and computational bottlenecks that limit classical ML models across MC domains, ranging from autonomous defense networks to energy grids.}  However, several fronts have to be addressed before realizing the potential of QAI in MC systems, including the lack of standardized QAI benchmarks, the fragility of current-day quantum hardware, and the interpretability gap between the certificable decision frameworks and quantum models.  Future developments in QAI for MC systems are likely to emerge from hybrid quantum-classical models optimized for near-term NISQ devices that integrate quantum error correction tailored for AI workloads.  A significant research direction in this field is the evolution of explainable QAI and quantum uncertainty to the level of verifiable standards in order to gain acceptance in safety-critical environments.  Finally, the convergence of intelligent decision systems and quantum computing represents a paradigm shift towards intelligent systems that are secure, self-adaptive, and reliable in the face of uncertainty.
\textcolor{black}{It is also important to acknowledge that QAI may remain impractical for production MC deployment for years due to - limitations of NISQ hardware, the gap between QAI capabilities and MC certification requirements, and the pace of quantum hardware maturation. The value of the present review lies in mapping the research landscape so that progress toward this long-term goal can be structured and measured.}
\label{concl}
\bibliographystyle{IEEEtran}
\bibliography{bibliography1.bib}

\vskip 10\baselineskip plus -1fil

\end{document}